\pdfoutput=1
\documentclass[11pt,a4paper]{article}
\usepackage[hyperref]{acl2021}
\usepackage{times}
\usepackage{url}            % simple URL typesetting
\usepackage{latexsym}
\usepackage{mathtools}

\usepackage{microtype}
\usepackage{booktabs} % For formal tables
\usepackage{multirow}
\usepackage{graphics}	
\usepackage{subfigure}
\usepackage{tabularx}
\usepackage[english]{babel}
\usepackage[utf8]{inputenc}
\usepackage{amsfonts}
\usepackage{graphicx}
\usepackage[colorinlistoftodos]{todonotes}
\usepackage{amsmath}
\usepackage[linesnumbered,ruled, vlined]{algorithm2e}
\usepackage{algorithmic}
\usepackage{bbm}

\newcommand{\kw}[1]{{\ensuremath {\mathsf{#1}}}\xspace}
\newcommand{\stitle}[1]{\vspace{0.75ex}\noindent{\bf #1}}

\newcommand{\eat}[1]{}

\usepackage{amsthm}

\newcommand{\mr}{\kw{MR}}
\newcommand{\reight}{\kw{R8}}
\newcommand{\rft}{\kw{R52}}
\newcommand{\ohsumed}{\kw{Ohsumed}}
\newcommand{\news}{\kw{20news}}
\newcommand{\imdb}{\kw{IMDB}}
\newcommand{\our}{\kw{DECA}}

\newcommand{\vl}{\mathbf{v}}
\newcommand{\cl}{\mathbf{c}}
\newcommand{\zl}{\mathbf{z}}
\newcommand{\hl}{\mathbf{h}}

\newcommand{\xl}{\mathbf{x}}
\newcommand{\uul}{\mathbf{U}}

\aclfinalcopy

\title{Unsupervised Document Embedding via Contrastive Augmentation}
\author{%
Dongsheng Luo$^1$\thanks{Equal Contribution}\quad Wei Cheng$^2$\footnotemark[1]\quad Jingchao Ni$^1$\quad Wenchao Yu$^2$\quad Xuchao Zhang$^2$\quad Bo Zong$^2$\\\quad \textbf{Yanchi Liu}$^2$\quad \textbf{Zhengzhang Chen}$^2$\quad\textbf{Dongjin Song}$^3$\quad\textbf{Haifeng Chen}$^2$\quad  \textbf{Xiang Zhang}$^1$\\
$^1$The Pennsylvania State University\\
$^2$NEC Labs America\\
$^3$
University of Connecticut\\
 \texttt{$^1$\{dul262,dux19,xzz89\}@psu.edu}\\ \texttt{$^2$\{weicheng,jni,wyu,bzong,yanchi,zchen,haifeng\}@nec-labs.com}\\
 \texttt{$^1$\{dongjin.song\}@uconn.edu}
}

\date{}

\begin{document}
\maketitle
\begin{abstract}
We present a contrasting learning approach with data augmentation techniques to learn document representations in an unsupervised manner. Inspired by recent contrastive self-supervised learning algorithms used for image and NLP pretraining, we hypothesize that high-quality document embedding should be invariant to diverse paraphrases that preserve the semantics of the original document. 
With different backbones and contrastive learning frameworks, our study reveals the enormous benefits of contrastive augmentation for document representation learning with two additional insights: 1) including data augmentation in a contrastive way can substantially improve the embedding quality in unsupervised document representation learning, and 2) in general, stochastic augmentations generated by simple word-level manipulation work much better than sentence-level and document-level ones. We plug our method into a classifier and compare it with a broad range of baseline methods on six benchmark datasets. Our method can decrease the classification error rate by up to 6.4\%  over the SOTA approaches on the document classification task, matching or even surpassing fully-supervised methods.
\end{abstract}

\section{Introduction}
\label{sec:intro}
Obtaining machine-understandable representations that capturing the semantics of documents have a huge impact on various natural language processing (NLP) tasks~\cite{bengio2003neural, collobert2008unified, mikolov2013distributed}, including sentiment analysis~\cite{dos2014deep}, information retrieval~\cite{clinchant2013aggregating},
word sense disambiguation~\cite{bhingardive2015unsupervised, hadiwinoto2019improved}, and machine translation~\cite{liu2019shared}. 
In this work, we study the unsupervised document embedding task which learns encoders that can efficiently encode documents into compact vectors to be used for different downstream tasks. 
In literature, researchers have proposed various models to obtain document embeddings~\cite{cheng2018treenet, clinchant2013aggregating, hill2016learning, logeswaran2018efficient, giorgi2020declutr}. 
% Deep learning based methods have shown the effectiveness in long text NLP tasks~\cite{lun2020multiple}. 
However, the quality of representations obtained by existing methods is significantly affected by the data scarcity problem~\cite{luque2019atalaya}, i.e., the lack of information in the low resource cases. 

\begin{table}[!t]
%\vspace{-0.1cm}
    \centering
    \begin{scriptsize}
        \begin{tabular}{c|c}
        \hline
        Augmentation Methods & Examples \\
        \hline
         Original & due to a \textbf{slighter} dispersion of roles \textbf{certainly}.\\
         WordNet & due to a \textbf{thin} dispersion of roles \textbf{surely}. \\
         PPDB  & ...due to a slighter \textbf{dispersal} of roles certainly.\\
         Antonym & due to a \textbf{not big} dispersion of roles certainly. \\
         Uninformative & due to a \textbf{thin} dispersion of roles certainly.\\
         Back-Translation & certainly due to a lighter distribution of roles.\\
            \hline
    \end{tabular}
    \end{scriptsize}
    \caption{Examples of different augmentations.}
    \label{tab:example}
\vspace{-0.5cm}    
\end{table}

Including more information is the key to the current challenge. Data augmentation is a popular technique that generates novel and realistic-looking training data with relatively lower quality than the original data points by applying a transformation to the example~\cite{devries2017improved}. Their quantities and diversities have shown the effectiveness on various learning algorithm in computer vision~\cite{chen2020simple}, speech~\cite{nguyen2020improving}, and NLP fields~\cite{sennrich2016improving, xie2020unsupervised, fadaee2017data, gao2019soft,wang2018switchout, wu2019conditional,shen2020simple,qu2020coda}. Focusing on the document representation learning, as shown in~\cite{marr1978representation, higgins2020representation}, only the impoverished projection of the reality is observed. Thus, the underlying semantics of a document is only partially expressed by the words that appear in it. Therefore, we can replace, delete, or insert some words in a document without changing its semantics or label information. With a sentence from a sentiment analysis dataset, IMDB, Table~\ref{tab:example} shows exemplar augmented sentences 
generated by different approaches that share the same semantic with the original example.

However, it is nontrivial to select appropriate augmentation techniques under unsupervised settings without knowledge of any label information. Even slight modifications may lead to substantial changes in their semantics, making the models more brittle~\cite{wang2018switchout}. Simply feeding more augmented texts to the learning model may lead to instability and sub-optimal embedding performance. Recently, contrastive learning attracts many researchers for its soaring performance in representation learning~\cite{chen2020simple, you2020graph, wu2018unsupervised,he2020momentum,xie2020unsupervised}. The successes of applying contrastive learning to push the state of the art forward in computer vision~\cite{chen2020simple} and semi-supervised learning~\cite{xie2020unsupervised} have shown significant promise to effectively incorporate diverse but noisy augmented data.
Inspired by the learning strategy that humans deploy - try to
find the commonalities between similarly expressed texts and contrast them with examples from other different texts, contrastive learning loss aims to maximizes the consistency under differently augmented views, enabling data-specific choices to inject the desired invariance~\cite{chen2020simple,you2020graph}. 
We hypothesize that a contrastive loss to incorporate stochastically augmented data can enforce the model to be insensitive to the noise in the augmented texts and hence smoother with respect to diverse paraphrases that preserve the semantics of the original document.
The resulting model well mitigates the data scarcity problem, especially in the low resource cases, and 
provides increased flexibility to choose different  augmentations, thus leads to better embedding performance.

% \textcolor{red}{---------------------------------update------------------------}

Our work systematically investigates the contrastive learning with different augmentations for unsupervised document representation learning to address the challenge of data scarcity. 
We adopt a simple but effective method, Doc2vecC~\cite{chen2017efficient}, and a pre-trained model, BERT~\cite{lan2019albert}, as backbones, respectively. Doc2vecC computes a document embedding by simply averaging the embeddings of all words in the document. 
BERT stacks Transformer layers, each consisting of a  self-attention sub-layer and a feed-forward sub-layer, to encode tokens in an input sequence~\cite{devlin2019bert}. The proposed framework has the advantage of being conceptually simple and easy to implement, requiring no specialized architectures, and can also be easily adapted to other sophisticated models~\cite{cheng2018treenet, conneau2017supervised, cozma2018automated, hill2016learning, kiros2015skip,beltagy2020longformer}.

We empirically show that 1) including  data augmentation in  a contrastive way can substantially improve the embedding quality in unsupervised document representation learning. 2) In general, word-level manipulation to generate realistic stochastic augmentation examples, such as synonym replacement,  works much better than augmentations in other granularity, such as sentence-level and document-level ones. This provides new perspectives of data augmentations that are different from previous findings under the supervised or semi-supervising settings~\cite{xie2020unsupervised,qu2020coda}. 3) Without fine-tuning with labels, recent Transformer-based models, such as BERT~\cite{lan2019albert,beltagy2020longformer} are not comparative with traditional methods for unsupervised document embedding, although they are privileged in sentence representation learning.   
We report a rich set of experiments characterizing the performance of our approach on a wide variety of benchmark datasets and in several different downstream tasks (linear classification and clustering). The results show that our method, \our (Document Embedding via Contrastive Augmentation) can significantly improve the quality of document embeddings with flexible choices of different data augmentations. It reduces the classification error rate by up to  6.4\% and relatively improves clustering performance by up to  7.6\% comparing to the second-best baselines. Surprisingly, in the classification task, our method can match or even surpass fully-supervised methods.

\vspace{-0.08cm}
\section{Related Work}
\vspace{-0.08cm}
\label{sec:relatedwork}
Bag-of-words (BoW) as well as its extensions, such as N-gram and TF-IDF, are the most commonly used methods to represent documents. In spite of their  simplicity, BoW based models are surprisingly effective for many NLP tasks~\cite{chen2017efficient}. To get low-dimensional representations, language model based methods~\cite{croft2003language,mikolov2010}, topic models~\cite{deerwester1990indexing, blei2003latent}, autoencoders~\cite{hill2016learning,gan2017learning}, distributed vector representations~\cite{le2014distributed, kiros2015skip} are further proposed.

Extending from Word2Vec~\cite{mikolov2013distributed}, Dov2vec learns a document embedding with context-word predictions~\cite{le2014distributed}. The document embedding matrix is kept in memory and is jointly optimized along with word embeddings. The model complexity is linear to the number of documents in the corpus, which hinders Doc2vec from being utilized in large scale datasets. Moreover, in the inductive setting, getting a representation for a new document is expensive for Doc2vec. Dov2vecC is then proposed to address these limitations by building document representation as the average of word embeddings~\cite{chen2017efficient}. 
Another thread of work builds document or sentence embeddings from pre-trained word embedding with two-stage pipelines~\cite{cheng2018treenet,conneau2017supervised,hill2016learning,kiros2015skip,kusner2015word,clinchant2013aggregating, arora2016simple,wu2018unsupervised,kiros2015skip,logeswaran2018efficient}. The method proposed in~\cite{arora2016simple} represents a sentence by a weighted average of pre-trained word embeddings generated from large scale corpus such as Wikipedia, fine-tuned with PCA/SVD.  WME~\cite{wu2018word} utilizes Word Mover’s
Distance to align semantically similar words when learning a semantics-preserving sentence representation from pre-trained word embeddings. 
% However, the cost scales with the size of training samples such that it is hard to be applied on large-scale dataset.
Denoised Autoencoder based methods are trained to reconstruct the original input data from the corrupted ones, which are generated with some noise functions~\cite{hill2016learning,gan2017learning}. These methods are able to capture the important factors of variation.  SDAE adopts the LSTM-based encoder-decoder architecture~\cite{hill2016learning} and CNN-LSTM architecture is utilized in~\cite{gan2017learning}.  However, the performances of these methods are substantially dominated by the quality and size of the applied corpus, which may lead to unsatisfactory representations in the low resource scenarios. 

Recently, Transformer-based models have achieved great success in various NLP tasks~\cite{vaswani2017attention, devlin2019bert, yang2019xlnet, beltagy2020longformer, zaheer2020big}.  BERT includes a special token [CLS] as the aggregated representation of the whole sequence, which is utilized in the downstream task. However, the maximum sequence length in BERT is 512, which limits BERT to be applied for long sequences, such as documents. To address this concern, Longformer, and BIGBIRD are proposed to adopt sparse attention mechanisms~\cite{beltagy2020longformer,zaheer2020big}. These sophisticated models are orthogonal to the proposed method and can be combined to further improve the quality of document embeddings.

\vspace{-0.06cm}
\section{Model}
\vspace{-0.06cm}
\label{sec:model}
% In this section, we present our method first formulate our task and then 
% To facilitate downstream tasks, a document embedding is required to compress useful features into a compact representation.
% It is not an easy task to learn discriminable features unsupervisedly since validation information is not accessible for training.
In this section, we first introduce some important notations and formulate the research problem. Then we introduce the framework of leveraging contrastive learning for unsupervised document embedding. Next, we briefly describe an efficient and effective instantiation with Doc2vecC~\cite{chen2017efficient} as the backbone. Last, we analyze different types of data augmentations used.

\subsection{Notations and Problem Formulation}
Notations we use throughout the paper are listed as follows.\vspace{-0.25cm}
\begin{itemize}
    \item $D_i$: the $i$-th document consisting of a sequence of words $w_i^1,w_i^2,...,w_i^{T_i}$, where $T_i$ is the length of $D_i$.\vspace{-0.35cm}
    \item $\mathcal{D}=\{D_1,D_2,...D_n\}$: a text corpus with $n=|\mathcal{D}|$ documents. \vspace{-0.3cm}
    \item $\mathcal{V}$: the vocabulary in the corpus $\mathcal{D}$, with the size $v=|\mathcal{V}|$.\vspace{-0.35cm}
    \item $\mathbf{x}_i\in \mathcal{R}^{v\times 1}$: the BoW representation vector of document $D_i$, Similar to one-hot coding, $x_{ij}=1$ iff word $j$ appears in document $D_i$.\vspace{-0.35cm}
    \item $\mathbf{h}_i\in \mathcal{R}^{d\times 1}$: the compact representation of document $D_i$, with $d$ as the dimensionality.\vspace{-0.35cm}
    \item $\tilde{D}_i$: a document generated by applying augmentations on $D_i$.\vspace{-0.35cm}
    \item $\tilde{\mathbf{x}}_i\in \mathcal{R}^{v\times 1},\tilde{\mathbf{h}}_i\in \mathcal{R}^{d\times 1}$: the BoW representation and compact representation of the augmented document $\tilde{D}_i$, respectively.\vspace{-0.25cm}
\end{itemize}
\stitle{Problem Formulation.} Our goal is to learn a function $f: \mathcal{D}\rightarrow \mathcal{R}^{d\times n}$ that maps a document $D_i$ to a compact representation with semantics preserved.

\begin{figure}
\vspace{-0.3cm}
    \centering
\setlength{\abovecaptionskip}{0.cm}
    \includegraphics[width=3in]{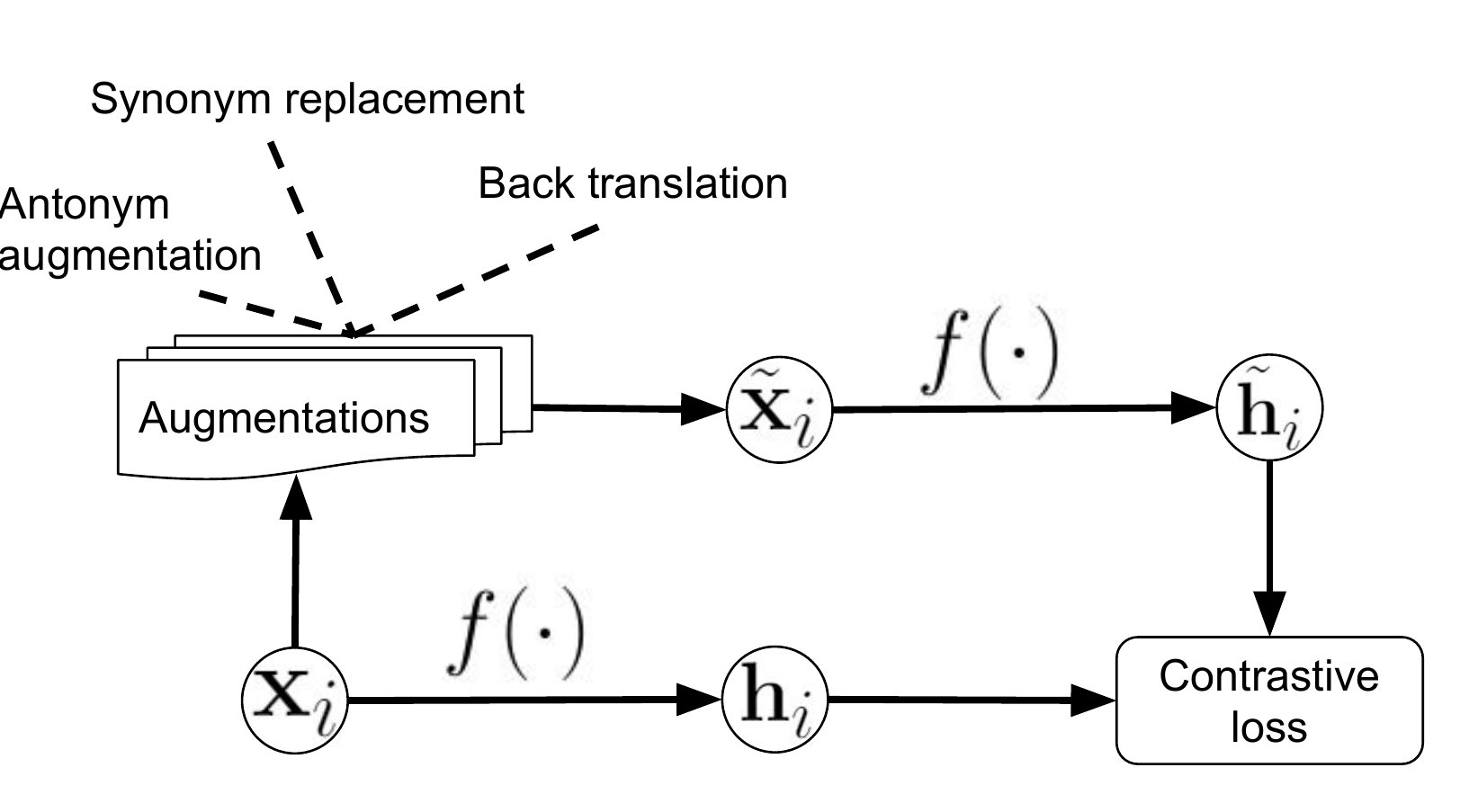}
    \caption{An illustration of the \our framework. $\mathbf{x}_i$ and $\tilde{\mathbf{x}}_i$ are the BoW representations of an input document $D_i$ and the corresponding augmented document $\tilde{D}_i$, respectively. $f(\cdot)$ is the encoder function to learn compact representations for documents. $\mathbf{h}_i$ and $\tilde{\mathbf{h}}_i$ are compact representations of $D_i$ and $\tilde{D}_i$, respectively.}
    \label{fig:framework}\vspace{-0.4cm}
\end{figure}

\subsection{Contrastive Learning Framework}
Document representation learning aims to obtain a low-dimensional embedding for a document that preserves its semantic meaning. Since the semantics of a document are not fully covered by the words that appear in it, data augmentations can be adopted to include more information by generating new documents that keep the same or similar semantics. Their diversity and quantity benefit various tasks in the fields of computer vision and speech~\cite{chen2020simple, nguyen2020improving}.  To take advantage of data augmentation, we propose a new model, referred to as \our (Document Embedding via Contrastive Augmentation)  for unsupervised document representation learning. The framework of \our is shown in Figure~\ref{fig:framework}. Pseudocode of the DECA algorithm is summarized in the Appendix~\ref{sec:code}. It consists of the following major components.

\stitle{Stochastic data augmentation module} generates a new augmented document $\tilde{D}_i$ of an input document $D_i$. We analyze five augmentation methods, including WordNet replacement, PPDB replacement, antonym replacement, uninformative word replacement, and back-translation. The details will be introduced in Section~\ref{sec:augmentation}.% $<\tilde{D}_i,D_i>$ is used as a positive pair in the contrastive learning.

\stitle{Document encoder}, denoted by $f:\mathcal{D}\rightarrow \mathcal{R}^{d\times n}$, computes a low-dimensional embedding of a document $D_i$ from its BoW presentation $\mathbf{x}_i$. In this study, we mainly utilize a simple but effective method, Doc2vecC~\cite{chen2017efficient}, as an instantiation. 
Doc2vecC compute the document embeddings as the mean of its word embeddings, motivated by the semantic meaning of linear operations on word embeddings calculated by Word2Vec~\cite{mikolov2013distributed}.
For example, vec(``man'') - vec(``woman'') $\approx$  vec(``uncle'') - vec(``aunt'')~\cite{mikolov2013linguistic}. Formally, 
\begin{equation}
\label{eq:embedding}
\setlength{\abovedisplayskip}{2pt}
\setlength{\belowdisplayskip}{2pt}
    \mathbf{h}_i \equiv f(D_i)= \frac{1}{T_i} \mathbf{U}\mathbf{x}_i,
\end{equation}
where $\mathbf{U}$ serves as the word embedding matrix. To optimize $\mathbf{U}$, Doc2vecC extends the Continuous Bag of Words Model (CBOW)~\cite{mikolov2013distributed} model by treating the document as a special token to the context and maximize the following probability for a target word $w^t$:
\begin{equation}
\label{eq:pp}
\setlength{\abovedisplayskip}{2pt}
\setlength{\belowdisplayskip}{2pt}
P(w^t|\cl^t, \xl)= \frac{\exp(\vl_{w^t}^\top (\uul\cl^t+\mathbf{h}))}{\sum_{w' \in \mathcal{V}} \exp(\vl_{w'}^\top \left(\uul\cl^t+\hl)\right)}
\end{equation}
Here, we remove subscript $i$ for simplicity. $\cl^t$ is the local context of the target word $w^t$ in the text window of document $D$. $\mathbf{V}$ is a learnable projection matrix. The element-wise loss function of Doc2vecC is
\begin{equation}
    \label{eq:doc2vecloss}
    \setlength{\abovedisplayskip}{2pt}
\setlength{\belowdisplayskip}{2pt}
    \ell_d^{(i)} = - \sum_{t=1}^{T_i} \log P(w^t_i|\cl^t_i, \xl_i).
\end{equation}
The sum up loss of an $N$ size mini-batch is $\ell_d = \sum_{i=1}^N \ell_{d}^{(i)} $.
We also adopt corruption and negative sampling when training the model~\cite{chen2017efficient}.
As a variant, we also include BERT~\cite{devlin2019bert} as another backbone. Specifically, we treat a document as a sentence and use [CLS] embedding as the document representation.

\stitle{Contrastive loss} is introduced as a regularizer, which is jointly optimized with the encoder loss $\ell_d$ to leaverage the augmented data for better embedding quality. The contrastive loss simply regularizes the embedding model to be invariant to diverse paraphrases that preserve the semantics of the original document. Encouraging consistency on the augmented examples can substantially improve the sample efficiency. Given a batch of $N$ documents $\{D_i\}_{i=1}^{N}$,  for each document $D_i$, an augmented document $\tilde{D}_i$ is generated by the stochastic data augmentation module.  Two state-of-the-art representative contrastive learning architectures, SimCLR~\cite{chen2020simple} and SimSaim~\cite{chen2020exploring} and are utilized. 

In the SimCLR framework, a pair of a document and its augmented document $<D_i, \tilde{D}_i>$ is treated as a positive pair and other $N-1$ pairs $\{<D_i,\tilde{D}_k>\}_{i\neq k},$ are considered as negative ones. The contrastive loss aims to identify $\tilde{D}_i$ out of the augmented documents in the batch for an input document $D_i$. Based upon NT-Xent loss~\cite{chen2020simple}, the sample-wise contrastive loss is defined by
\begin{equation}
\label{eq:constrativeloss}
\setlength{\abovedisplayskip}{1pt}
\setlength{\belowdisplayskip}{1pt}
    \ell_{c}^{(i)} = -\log \frac{\exp(\text{cos}(\hl_i, \tilde{\hl}_i)/\tau)}{\sum_{k=1}^{N} \mathbb{I}_{[{k \neq i}]}\exp(\text{cos}(\hl_i, \tilde{\hl}_k)/\tau)},
\end{equation}
where $\hl_i$ and $\tilde{\hl}_i$ are document embeddings calculated by Eq.~(\ref{eq:embedding}); $\text{cos}(\cdot,\cdot)$ denotes the cosine similarity between vectors; $\tau$ is the temperature parameter.

Considering that SimCLR relays large batch sizes to retrieve the negative pairs, SimSaim learns representations without adopting negative pairs. Instead, it adopts a predictor and stop gradient to avoid collapsed solutions~\cite{chen2020exploring}. Within the SimSaim framework, a prediction MLP with Batch Normalization is first applied to get output vectors: $\mathbf{z}_i=f(\hl_i)$ and $\tilde{\mathbf{z}}_i = f(\tilde{\hl}_i)$. The objective is to minimize the negative cosine similarity between $\hl_i$, $\tilde{\zl}_i$ and $\tilde{\hl}_i$, $\zl_i$. Stop gradient is utilized in one way to avoid the collapsed solution. Specifically, the loss function is defined by 
\begin{equation}
\label{eq:simsaimcl}
\setlength{\abovedisplayskip}{1pt}
\setlength{\belowdisplayskip}{2pt}
    \ell_{c}^{(i)} = \frac{1}{2}\mathcal{D}(\hl_i,\text{stopgrad}(\tilde{\zl}_i))+\frac{1}{2}\mathcal{D}(\tilde{\hl}_i,\text{stopgrad}(\zl_i)).
\end{equation}
The function $\mathcal{D}(\cdot,\cdot)$ is the negative cosine similarity:
$$\mathcal{D}(\mathbf{x},\mathbf{y}) = -\frac{\mathbf{x}}{||\mathbf{x}||_2}\cdot \frac{\mathbf{y}}{||\mathbf{y}||_2}.$$
Then the sum up loss of a batch is $\ell_c = \sum_{i=1}^N \ell_{c}^{(i)}$.

With the contrastive loss $\ell_c$ as a regularization term, the objective function of \our with mini-batch is to minimize the following loss function:
\begin{equation}
    \label{eq:objective}
    \setlength{\abovedisplayskip}{1pt}
\setlength{\belowdisplayskip}{1pt}
    \ell = \ell_d + \lambda \ell_c = \sum_{i=1}^N [-\sum_{t=1}^{T_i} \log P(w^t_i|\cl^t_i, \xl_i)+\lambda \ell_c^{i}],
\end{equation}
where $\lambda$ is a hyper-parameter to get a trade-off between two losses. When BERT is adopted as the backbone, we directly fine-tune it with the contrastive loss $\ell_c$.

\subsection{Augmentation Strategies}
\label{sec:augmentation}
Generating realistic augmented examples that preserve the semantics of original documents in an efficient way is non-trivial. In this section, we introduce five augmentation strategies, WordNet replacement~\cite{miller1995wordnet}, PPDB replacement~\cite{ganitkevitch2013ppdb}, negative-antonym-based augmentation~\cite{miller1995wordnet}, uninformative word replacement~\cite{xie2020unsupervised}, and back-translation~\cite{xie2020unsupervised,qu2020coda}. 
These methods are non-parametric and dataset agnostic that can be directly applied to any text corpus.

\stitle{Stochastic Thesaurus-based Substitution}.
We consider three kinds of word-level thesaurus-based substitutions. With these augmentations, our goal is to paraphrase the input document by replacing words based on synonyms, antonym with a negative prefix, or their frequencies, and at the same time, keep its semantics.\vspace{-0.3cm}
\begin{itemize}
    \item \textit{Synonyms Replacement.} For each word, we first extract a set of replacement candidates using WordNet synsets~\cite{miller1995wordnet} and filter out the ones out of vocabulary or with low frequencies. For efficient computation, the original word is also included in its synonym set.  To generate an augmented document, for each word, we randomly select a word in the set of its replacement candidates. Similarly, we also adopt PPDB~\cite{ganitkevitch2013ppdb} to generate replacement candidates as another augmentation method.\vspace{-0.3cm}
    \item \textit{Negative-Antonym-based Replacement.} For this method, we replace an adjective or a verb by its antonym with a negative prefix, like ``not''. For example, a review from the IMDB dataset ``...offers an exceptionally \textbf{strong} characterization'' is augmented to ``... offers an exceptionally \textbf{not impotent} characterization'' by this negative-antonym-based replacement. We extract the candidate sets using WordNet synsets and words POS tags~\cite{miller1995wordnet}. Out-of-vocabulary ones are filtered out.\vspace{-0.3cm}
    \item \textit{Uninformative Word Replacement.} As shown in~\cite{xie2020unsupervised}, for topic-related datasets, such as DBPedia, keywords are more informative in determining the topic. Besides, different from  the supervised and semi-supervised settings, infrequent words have subtle effects on the quality of the whole document embedding~\cite{chen2019self,chen2017efficient}. Thus, previous work removes words that appear less than 10 times with a special ``UNK'' token~\cite{chen2019self} or randomly replaces  them with other uninformative words. Unlike these methods, we substitute these low frequent words with synonyms with high frequencies~\cite{xie2020unsupervised}. This strategy is similar to the ``Word replacing with TF-IDF'' in a recent work UDA~\cite{xie2020unsupervised}.\vspace{-0.3cm}
\end{itemize}

\stitle{Back-Translation} first translates a document $D$ from the original language (English in this study) to another language, like German and French, to get $D'$. Then, the document $D'$ is translated back to the original language as the augmented document $\tilde{D}$~\cite{xie2020unsupervised, regina2020text}. Comparing to word level thesaurus-based substitution, as pointed out in~\cite{xie2020unsupervised},  document-level back-translation can generate paraphrases with high diversities while preserving the semantics. Back-translation has been used in many related NLP tasks, such as question answering and text classification to boost the accuracy performance~\cite{xie2020unsupervised, regina2020text}. 
\vspace{-0.08cm}
\section{Experiments}
\vspace{-0.08cm}
\label{sec:exp}
We conduct an extensive set of experiments to characterize the performance of our approach on a wide variety of benchmark datasets and in two downstream tasks--linear classification and clustering. Additional experiments are in Appendix~\ref{sec:appexp}.
% Our code is available at \url{https://github.com/anonymous-DECA1/DECA}

\vspace{-0.08cm}
\subsection{Setup}
\vspace{-0.08cm}

\stitle{Datasets.} 
A wide range of document corpora, including sentiment analysis (\mr, \imdb), news classification (\reight, \rft,\news) , and medical literature (\ohsumed), are adopted.  Detailed descriptions are given in Appendix~\ref{sec:dataset}.  When evaluating embeddings with text classification, we adopt the training/test split as in~\cite{yao2019graph, maas2011learning}.

\stitle{Baselines.}
We consider two backbones, Doc2vecC~\cite{chen2017efficient} and BERT~\cite{devlin2019bert}, abbreviated by DC and BT, and two contrastive learning architectures, SimCLR~\cite{chen2020simple} and SimSaim~\cite{chen2020exploring}, abbreviated by SC and SS. Recent pre-trained models for long sequences, such as Longformer~\cite{beltagy2020longformer},  are not adopted as the backbone due to the memory limitation. With 11GB memory, With Longformer as the backbone, \our encounters out-of-memory error even if the batch size is set to 1. The following state-of-the-art approaches are selected for comparison. 
\begin{itemize}
    % \item Bag-of-Word
    \item \textbf{TF-IDF} is the BoW model with term frequency inverse document frequency as the word weights.\vspace{-0.2cm}
    \item \textbf{LDA}~\cite{blei2003latent} is a generative model that gives a mix of topics for each document.
    \item \textbf{DEA}~\cite{hill2016learning} (denoised autoencoders) learns low-dimensional representations by minimizing the reconstruction error.\vspace{-0.2cm}
    \item \textbf{W2V-AVG}~\cite{mikolov2013distributed} first applies wor2vec to learn word embeddings. A document embedding is calculated by averaging embeddings of words appear in it. \vspace{-0.2cm}
    \item \textbf{PV-DBOW}~\cite{le2014distributed} adds a paragraph (document) vector to the word2vec model~\cite{mikolov2013distributed}. \vspace{-0.2cm}
    \item \textbf{PV-DM}~\cite{le2014distributed} is a distributed memory model of paragraph (document) vectors.\vspace{-0.2cm}
    \item \textbf{Skip-Thought}~\cite{kiros2015skip} is an RNN-based encoder.\vspace{-0.2cm}
    \item \textbf{Doc2vecC}~\cite{chen2017efficient} treats document embedding as the mean of its word embeddings and includes corruption during training.\vspace{-0.2cm}
    \item \textbf{SDDE}~\cite{chen2019self} includes a discriminator to capture inter-document relationship.\vspace{-0.2cm}
    \item \textbf{BERT}~\cite{devlin2019bert} is a Transformer-based language representation model. The ``[CLS]'' embedding is used as the document embedding. We use a pre-trained BERT-base model with 12 layers and 768 attention heads. A BERT model fined-tuned with masked language modeling on the target dataset is also included, denoted by \textbf{BERT-mlm}. We also include another baseline \textbf{BERT-ST} that uses BERT to learn sentence embeddings and then gets document embeddings by summing their sentence embeddings. 
    \vspace{-0.2cm}
    \item \textbf{Longformer}~\cite{beltagy2020longformer} extends BERT~\cite{devlin2019bert} by adopting sparse attentions in Transformers to enable long-sentence usage. We utilize a pre-trained model with 12 layers, 768 hidden, 12 heads and 149M parameters. \textbf{LF-mlm} is fine-tuned with  masked language modeling. \vspace{-0.2cm}
\end{itemize}

For classification evaluation, we also include a representative fully supervised method, TextGCN~\cite{yao2019graph}, which adopts the graph convolutional network to learn document embeddings by creating a word-document graph.

\begin{table}[h!]
\centering
\begin{small}
\begin{tabular}{c|c|c|c|c}
\hline
Dataset      &Win. size        &\#Neg.         &Aug. type           &\#Sample \\
\hline
\reight           & 6        &5        &Antonym            &5\\
\rft          & 10       &5        &Antonym            &10\\
\mr           & 10        &5        &WordNet            &5\\
\ohsumed      & 10        &7        &WordNet            &7\\
\news       & 8        &5        &WordNet            &5\\
\imdb         & 4        &5        &PPDB           &15\\
\hline
\end{tabular}
\caption{Hyperparameters used in experiments. Win. size (window size), \#Neg. (number of words in the negative sampling) and \#Sample (number of words in each sampled document) are hyper-parameters in the backbone Doc2vecC. }
\label{tab:setting}
\end{small}\vspace{-0.3cm}
\end{table}

\stitle{Settings.} The embedding dimensionality is set to 100, except for Transformer-based models, whose output dimension is 768.  For each dataset, we first use all documents to learn an embedding for each one. Then, these document embeddings will be evaluated with two downstream tasks, linear classification and clustering. For retailed settings, please refer to Appendix~\ref{sec:settings}.

\subsection{Comparison with Classification}

\setlength{\tabcolsep}{0.14em}
\begin{table}[t!]
\centering
\vspace{0.2cm}
\begin{small}
\begin{tabular}{c|c|c|c|c|c|c}
\hline
Method      &\reight    &\rft    &\mr       &\ohsumed    &\news    &\imdb \\
\hline
TF-IDF      &5.2        &12.4    &25.4     &44.1      &17.2   &11.7\\
LDA      &7.0        &17.5    &46.9     &61.9      &33.9   &16.5\\
DEA         &4.8        &12.2    &29.7     &46.6      &20.9   &13.9 \\
W2V-AVG     &5.8        &15.3    &43.4     &65.5      &43.5    &12.7\\
PV-DBOW     &8.7        &18.4    &31.8     &54.4      &32.2   &13.1 \\
PV-DM       &12.7       &22.1    &43.8     &65.6      &32.6    &13.3\\
Skip-Thought&--       &--       & --      &--       &45.9   &13.5\\
% WME         &----       &----       & ---       &----       &----    &----\\
SDDE         &--       &--       & --       &--       &24.4    &10.2\\
BERT         &5.1       &11.2    &18.7        &51.0       &32.9    &10.8\\
BERT-ST      &5.1       &11.2    &18.7        &51.9       &41.1    &11.1\\
BERT-mlm     &4.1        &7.7    &18.9        &45.8       &26.5    &12.7\\
Longformer   &5.1       &17.9     &17.7          & 55.2  &29.6    &11.2\\
LF-mlm       &3.5         &12.3     &\textbf{15.9}          &49.2  &30.2    &14.3\\
Doc2vecC    &3.0      &9.6      &29.3      &45.5      &19.2   &10.5 \\
% \hline
\underline{TextGCN}   &2.9      &\textbf{6.4}    &23.3     &\textbf{31.5}      &\textbf{13.7}   &-- \\
\hline
\our (DC+SC) & \textbf{2.4} &\textbf{6.4} &28.1 	& \textbf{39.1}	  &\textbf{14.9} &\textbf{9.8}	\\
\our (DC+SS) &2.8  &9.2 	& 29.1	  &41.6 &17.2&10.3\\
\our (BT+SC) &6.6 &15.3 	&20.9	 &62.8&61.1 &42.3\\
\our (BT+SS) &4.9 &14.3 	&21.0  &57.3&41.6&50.0\\
\hline
\end{tabular}
\caption{Performance comparison with classification error rate}
\label{tab:ac}
\end{small}
 \vspace{-0.3cm}
\end{table}
We first adopt classification as the down-stream task to evaluate the quality of embeddings calculated by each method.
Linear classification is a standard benchmark for evaluating the quality of unsupervised document representations. Logistic regression~\cite{scikit-learn} is adopted as the classifier and the testing error rate is used as the evaluation metric. Visualization of embeddings on 20news dataset is illustrated in the Appendix~\ref{sec:vis}.

Table 2 presents accuracy performances on the test set of each method. 
Our method (DC+SC) achieves the best performances in five out of six datasets, verifying the effectiveness of \our on unsupervised document embedding. The only exception happens in the \mr dataset, in which each document includes only one single sentence. In that case, the pre-trained models like BERT and Longformer can better encode the context and achieve better performance. To achieve an efficient \textit{document} rather than \textit{sentence embedding}, we adopt a simple method, Doc2vecC as the backbone in DC+SC. However, the performance of it is much worse than Transformer-based methods in the \mr dataset, therefore, \our (DC+SC) is not comparable to BERT and Longformer in this dataset. 

We also have the following observations from the table. 1) Simple methods such as TF-IDF and Doc2vecC achieve competitive performances in most datasets. This observation is also aligned with~\cite{yao2019graph}. 2) Comparing \our (DC+SC) with its backbone Dov2vecC, including contrastive augmentations reduces the absolute classification error rates by  0.6\%, 3.2\%, 1.2\%, 6.4\%, 4.3\%, and 0.7\% on \mr, \reight, \rft, \ohsumed, \news and \imdb, respectively.  3) The comparison between DC+SC and DC+SS shows that with Doc2vecC as the backbone,  SimCLR is a better choice as the contrastive framework. The reason is that Doc2vecC is a simple and efficient backbone. The large batch size ensures qualities of negative samples for contrastive learning.  The observation is consistent with the original paper~\cite{chen2020simple}. 4) A recent work demonstrates the effectiveness of contrastive learning for pre-training BERT on sentence representation learning~\cite{wu2020clear}. However, for the unsupervised document representation learning, fine-tuning with contrastive losses decreases the performances of BERT because of the memory limitation. With an 11GB GPU, the batch size can only be set up to 2. SimCLR framework fails due to the insufficient number of negative pairs~\cite{chen2020simple}.  As shown in SimSaim paper, Batch Normalization is a necessary component~\cite{chen2020exploring}, which works with sufficient large batch sizes. At the same time, SimSaim framework, in general, achieves better results than SimCLR framework with small batch sizes, which is aligned with the observation in the SimSaim paper~\cite{chen2020exploring}. Deep insight into BERT with contrastive loss w.r.t batch size is provided in Appendix~\ref{sec:batchsize} . The second reason is that BERT is pre-trained on large scale unlabelled data. Data augmentations such as stochastic thesaurus-based substitution and back translation, on the small scale labeled data provide insufficient guidance considering the huge size of parameters.  5) Skip-Thought~\cite{kiros2015skip} and SDDE~\cite{chen2019self} rely on the sentence-document relationship to encode the text, while in \mr, \reight, \rft, and \ohsumed, there are only one or a few sentences in each document. Thus, these two methods are not applicable to these datasets. 6) With contrastive augmentation, \our matches or even surpasses a SOTA fully-supervised method TextGCN~\cite{yao2019graph}. The result of TextGCN on \imdb is missing because it is infeasible to handle such a large dataset.

The reason why \our works well is twofold. First, data augmentations used in \our generate new documents with relatively low qualities, enriching the diversity of the text dataset, which is the key to address the low resource problem.  Second, in the contrastive learning framework, \our is more robust to noise introduced in the augmented texts, which equips \our with more flexibility to choose different augmentation methods and leads to embeddings with higher quality. Deeper insights of \our will be given in Section~\ref{sec:deep}.

\setlength{\tabcolsep}{0.4em}
\subsection{Comparison with Clustering}
%\vspace{-0.08cm}
\label{sec:clusteing}

Clustering aims to find groups of samples without label information. In this section, we adopt clustering as another downstream task to evaluate the quality of embeddings generated by different methods. We utilize the $k$-means to conduct clustering and adopt the normalized mutual information (NMI) as the evaluation metric. Two sentence-based methods and a fully supervised method are not included in this set of experiments. Two sentiment analysis datasets \mr and \imdb are not used as their label information is unrelated to the text topics and thus not suitable for clustering evaluation.  

The comparison results are shown in Table~\ref{tab:clustering}. Our method \our achieves the best performances in all four selected benchmark datasets. The performance gain over the second-best baseline is up to 7.6\% in terms of NMI.

\begin{table}[t!]
\vspace{0.2cm}
\centering
\begin{small}
\begin{tabular}{c|c|c|c|c}
\hline
Method      &\reight     &\rft      &\ohsumed    &\news \\
\hline
TF-IDF      &0.358       &0.491     &0.223       &0.340\\
LDA      &0.360       &0.416     &0.118       &0.410\\
DEA         &0.496       & 0.521    &0.213       &0.547\\
W2V-AVG     &0.511       &0.474     &0.080       &0.208\\
PV-DBOW     &0.385       &0.421     &0.107       &0.443\\
PV-DM       &0.350       &0.353     &0.103       &0.302\\
BERT        &0.304       & 0.358   &0.086        &0.205\\
BERT-ST     &0.304       & 0.358   &0.047        &0.028\\
BERT-mlm    &0.355       &0.447   &0.089         &0.253\\
Longformer  &0.383       & 0.449    &0.071       &0.214\\
LF-mlm      &0.448          &0.515    &0.144       &0.227\\
Doc2vecC    &0.517      & 0.529     &0.182       &0.589\\
\hline
\our (DC+SC) & \textbf{0.592} &\textbf{0.533} 	& \textbf{0.237}	  &\textbf{0.634}\\
\our (DC+SS) & 0.541  &0.532 	&0.225	  &0.601\\
\our (BT+SC) &0.205 &0.255 	& 0.033	  &0.032\\
\our (BT+SS) &0.157 &0.225 	&0.048 &0.035\\

\hline
\end{tabular}
\caption{Performance comparison with clustering NMI}
\label{tab:clustering}
\end{small}
\vspace{-0.35cm}
\end{table}

%\vspace{-0.09cm}
\subsection{Insights into the Model}
%\vspace{-0.09cm}
\label{sec:deep}
\setlength{\tabcolsep}{0.2em}
\begin{table}[h!]
\vspace{-0.2cm}
\centering
\begin{scriptsize}
\begin{tabular}{c|c|c|c|c|c}
\hline
Method           & Wordnet     &PPDB         & Antonyms    & Uninformative    &Back Trans.  \\
\hline
\our             &2.74        &2.42          &2.38     & 2.96   &2.97    \\
-Contrastive     & 3.02       &2.92          & 2.92    &2.96    & 3.15         \\
-Dictionary      &3.83        &3.15          & 2.97    &-       &- \\
\hline
\end{tabular}
\caption{Classification error rates of \our and its non-contrastive variants on \reight. (Doc2vecC is 3.0.)}
\label{tab:insight}
\end{scriptsize}
\vspace{-0.3cm}
\end{table}
In this part, we provide experiments to show deep insights into our method. We adopt Doc2vecC as the backbone, SimCLR as the contrastive framework and use the \reight dataset as an example. We first check the performance gains obtained by including different types of augmentations. From the second row in Table~\ref{tab:insight}, we observe that word-level thesaurus-based substitutions consistently outperform the document-level one, back-translation. As discussed in~\cite{xie2020unsupervised,sennrich2016improving}, back-translation can generate diverse augmentations with diversity and benefit supervised or semi-supervised tasks in NLP.  We found that for unsupervised document representation learning, the simple word-level replacement works much better. 

To verify the effectiveness of contrastive learning, we compare \our with another usage form of data augmentation that randomly samples an augmented document and appends it to the end of the original document. Then we apply Doc2vecC on this merged corpus to learn document embeddings. Results are shown in Table~\ref{tab:insight} (row ``-Contrastive'').  With back-translation as the augmentation, the performance of Doc2vec even drops, because back-translation generates documents with lower quality and includes noise. With the contrastive learning framework, \our can consistently boost the performance, demonstrating the robustness of \our. We have similar observations in other datasets. For example, for \news we decrease the error rate by 1.6\% when including data augmentations, and the gain is 4.3\% when using contrastive augmentations.

We empirically evaluate the effectiveness of including the dictionary constraint that only words inside the vocabulary are considered as replacement candidates. The results of \our without this constraint and contrastive loss are shown in the last row in Table~\ref{tab:insight}. The comparison between this variant and ``-Contrastive'' shows that dictionary constraint plays a vital role in the unsupervised document representation learning. This also explains why back-translation does not benefit our task much, since it generates highly diverse paraphrases without the inside dictionary constraint. This study provides new perspectives of data augmentations that are different from previous findings under supervised or semi-supervising settings~\cite{xie2020unsupervised,qu2020coda}.

\vspace{-0.1cm}
\section{Conclusion}
\vspace{-0.1cm}
\label{sec:conclusion}
In this paper, we have studied the unsupervised document embedding problem and have proposed a contrasting learning approach with data augmentations. We have shown that contrastive augmentation is an effective way that substantially improves the quality of document embeddings. Besides, we also found that word-level manipulations work much better than sentence-level or document-level ones. As shown in the experiments, our method can generate high-quality embeddings and benefit various downstream tasks.

\clearpage
\section*{Broader Impact Statement}

Over the past years, Transformer-based models have been extensively studied and used for different NLP tasks. 
The majority of efforts are devoted to supervised learning or unsupervised pre-trained.  Instead of heading towards that direction,
this work explores the topic of unsupervised document representation learning, which has wide applications in various fields. We find that without any label information, traditional methods, such as Doc2vecC are more effective and efficient than recent transformer-based models.  Our results may invoke more discussions on the usage of large scale language models.

\bibliography{ref}
\bibliographystyle{acl_natbib}

\clearpage
\appendix

\section{Supplementary Materials}
\label{sec:appendix}

\subsection{Introduction}
This document is the Supplementary Materials for Unsupervised Document Embedding via Contrastive Augmentation.
It contains supportive materials for the work that are also important but unable to be completely covered in the main transcript due to the page limits. 
% These content involves the Section~\ref{sec:method} for the proposed methods and Section~\ref{sec:experiments} for experiments.

\subsection{Methods}
Supplementary information regarding the proposed methods provided in this section helps illustrate 
\subsubsection{Pseudocode of DECA}
\label{sec:code}
Algorithm~\ref{alg:framework} summarizes the proposed method.
\begin{algorithm}
\small{
    \centering
    \caption{\our (Document Embedding via Contrastive Augmentation)}
    \label{alg:framework}
\begin{algorithmic}[1]
    \STATE \textbf{input:} document set $\mathcal{D}$, vocabulary $\mathcal{V}$, embedding size $d$, mini-batch size $N$, hyper-parameters $\lambda, \tau$, data augmentation function $g(\cdot)$,  document encoder $f(\cdot)$
    \STATE \textbf{output:} document embeddings $\{\hl_i\}_{i=1}^{n}$
    \STATE Initialize word embedding matrix $\mathbf{U}$ and projection matrix $\mathbf{V}$.
    \FOR{sampled mini-batch $\{D_i\}_{i=1}^N$}
    \STATE \textbf{for all} $i\in \{1, \ldots, N\}$ \textbf{do}
        \STATE $~~~~$Generate an augmented document $\tilde{D}_i \leftarrow g(D_i)$
        % \STATE $~~~~$get BoW representations $\mathbf{x}_i$, $\tilde{\mathbf{x}}_i$ of documents $D_i$ and $\tilde{D}_i$, respectively.
        \STATE $~~~~$$\hl_i \leftarrow f_{\mathbf{U},\mathbf{V}}(D_i)$  
        \STATE $~~~~$$\tilde{\hl}_i \leftarrow f_{\mathbf{U},\mathbf{V}}(\tilde{D}_i)$
        \STATE $~~~~$Compute element-wise doc2vecC loss $\ell_d^{(i)}$ with Eq.~(\ref{eq:doc2vecloss}).
        \STATE $~~~~$Compute element-wise contrastive loss $\ell_c^{(i)}$ with Eq.~(\ref{eq:constrativeloss})
    \STATE \textbf{end for}
    \STATE Compute the batch loss $\ell=\sum_{i=1}^{N} [\ell_d^{(i)}+\lambda\ell_c^{(i)}]$
    \STATE Update parameters $\mathbf{U},\mathbf{V}$ with backpropagation.
    \ENDFOR
    \STATE \textbf{return} document embeddings $\{\hl_i\}_{i=1}^{n}$ computed by Eq. (\ref{eq:embedding}).
\end{algorithmic}
}
\end{algorithm}
\label{subsubsec:pseudocode}

\subsection{Experiments}
\label{sec:appexp}
This section introduces the detailed experimental set up and additional experiments.

\subsubsection{Datasets}
\label{sec:dataset}
\begin{table}[h!]
\centering
\begin{small}
\begin{tabular}{c|c|c|c|c}
\hline
Dataset      &\#Docs        &\#Training   &\#Test           &\#classes    \\
\hline
\reight           & 7,674        &5,485        &2,189            &8\\
\rft          & 9,100        &6,532        &2,568            &52\\
\ohsumed      & 7,400        &3,357        &4,043            &23\\
\mr           &10,062        &7,108        &3,554            &2\\
\news       &18,846        &11,314       &7,532            &20\\
\imdb         &100,000       &25,000       &25,000           &2\\
\hline
\end{tabular}
\caption{Statistics of datasets}
\label{tab:statistics}
\end{small}\vspace{-0.5cm}
\end{table}

\begin{itemize}
    \item \reight and \rft are two subsets drawn from the Reuters 21578 dataset\footnote{https://www.cs.umb.edu/~smimarog/textmining/datasets/}, a collection of documents on Reuters newswire in 1987. R8 has 7,674 documents from 8 categories, R52 has 9,100 documents from 52 categories. \vspace{-0.1cm}
    \item \mr is a movie review dataset for sentiment analysis, with 5,331 positive reviews and 5,331 negative reviews~\cite{pang2005seeing}.\vspace{-0.1cm}
    \item \ohsumed is from MEDLINE, an important database of medical literature. It consists of 7,400 documents from 23 categories.\vspace{-0.1cm}
    \item The \news dataset\footnote{http://qwone.com/7Ejason/20Newsgroups/} contains 18,846 documents of  20 different classes. \vspace{-0.1cm}
    \item The \imdb is a dataset for binary sentiment analysis, consisting of 100,000 movie reviews. 50,000 out of them are labeled.
\end{itemize}
The statistics of the used datasets are summarized in Table~\ref{tab:statistics}.

\subsubsection{Detailed settings}
\label{sec:settings}
We adopt the Adam method for optimization. Dropout ratio is set to 0.3.  When SimSaim is adopt as the contrastive loss, we adopt a two layer MLPs with 64 hidden. Batch Normalization are adopted. When Doc2vecC is adopted as the backbone, the batch size is set to 4,096. 
% $\lambda$ and $\tau$ are set to 0.1 and 1.0, respectively. 
$\tau$ is set to 1.0. 
Learning rate is set to 1e-3. The settings of other important hyper-parameters are summarized in
Table~\ref{tab:setting}. We adopt the pretrained BERT model\footnote{\url{https://huggingface.co/bert-base-uncased}} with 12-layer, 768-hidden, 12-heads, and 110M parameters as the backbone. The maximum length of input sequence is set to 512 by default. Due to the memory limitation, batch size is set to 2.  We set learning rate to 3e-6 and pretrain for 1 epoch with the contrastive loss.

\subsubsection{Visualization}
\label{sec:vis}
\begin{figure*}[h]
\vspace{-0.5cm}
    \centering
    \setlength{\abovecaptionskip}{0.cm}
    \subfigure[TF-IDF]{\includegraphics[width=1.8in,height=1.3in]{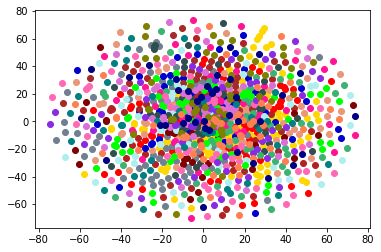}\label{fig:vis:tfidf}}
    \subfigure[DEA]{\includegraphics[width=1.8in,height=1.293in]{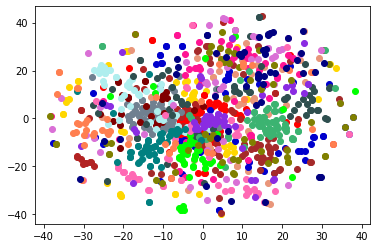}\label{fig:vis:dea}}
    % \subfigure[W2V-AVG]{\includegraphics[width=1.8in,height=1.3in]{figures/exp/word2vec.png}\label{fig:vis:word2vec}}
    \subfigure[PV-DBOW]{\includegraphics[width=1.8in,height=1.3in]{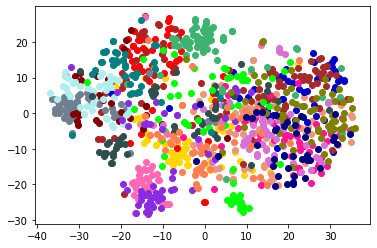}\label{fig:vis:PV-DBOW}}
    % \subfigure[PV-DM]{\includegraphics[width=1.8in,height=1.3in]{figures/exp/PV-DM.png}\label{fig:vis:PV-DM}}
    \subfigure[Doc2vecC]{\includegraphics[width=1.8in,height=1.3in]{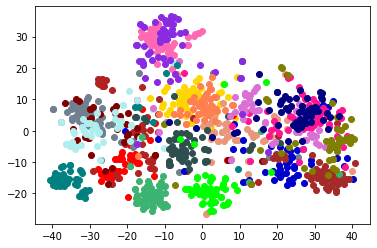}\label{fig:vis:Doc2vecC}}
    \subfigure[BERT]{\includegraphics[width=1.8in,height=1.3in]{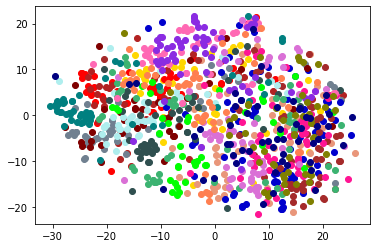}\label{fig:vis:bert}}
    \subfigure[\our]{\includegraphics[width=1.8in,height=1.3in]{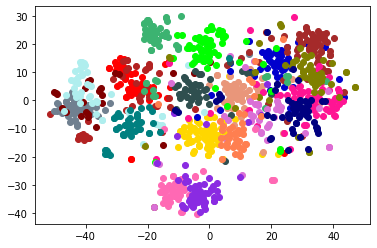}\label{fig:vis:deca}}
    \caption{Visualization of embeddings of different methods on \news dataset.}
    \label{fig:vis}
    \vspace{-0.45cm}
\end{figure*}
To better understand the difference between the compared methods, we randomly select 1,000 documents from the \news dataset and use t-SNE to project the document embeddings to a 2D space for visualization (Figure~\ref{fig:vis}). Different colors represent 20 different clusters.  We adopt \our (DC+SC) and select other five representative methods in this section.

In general, the visualization results shown in the figures are aligned with classification error rates. All samples in Figure~\ref{fig:vis:tfidf} mix with each other because of the high dimensionality of the original TF-IDF representations. This demonstrates the necessity of conducting dimensional reduction. Among the selected baselines, the visualization result of Doc2vecC is with clear margins between clusters. By including contrastive augmentation, our method \our further improves the quality of document embeddings. For example, pink and purple nodes are well separated in Figure~\ref{fig:vis:deca}, while mixed together in Figure~\ref{fig:vis:Doc2vecC}.

\subsubsection{The Effects of Batch Size}
\label{sec:batchsize}
We have experimentally shown that contrastive losses impair the performance of BERT in Table~\ref{tab:ac}. One potential reason is that the recent contrastive frameworks require large batch sizes (larger than 256). However, with BERT as the backbone, \our can only set the batch size up to 2 due to the memory limitation.  
Since each document in the \mr dataset contains only one sentence, we adopt \mr in this part and limit the maximum of the input sequence to 64 to enable larger batch sizes. We range the batch size from 2 to 8 and show the results of \our (BT+SC) and \our (BT+SS) in Table~\ref{tab:batchsize}.  The table shows that both methods benefit from large batch sizes. 

\begin{table}[h!]
\centering
\begin{small}
\begin{tabular}{c|c|c|c}
\hline
Batch Size           &2     &4         &8  \\
\hline
\our (DC+SC)     &21.69        &21.94          & 19.61\\
\our (DC+SS)     &20.99  & 20.65 & 19.31       \\
\hline
\end{tabular}
\caption{The effects of batch size on \our with BERT as the backbone}
\label{tab:batchsize}
\end{small}
%\vspace{-0.5cm}
\end{table}

\subsubsection{Parameter Sensitivity Study}
\begin{figure}
    \centering
    \includegraphics[width=3in]{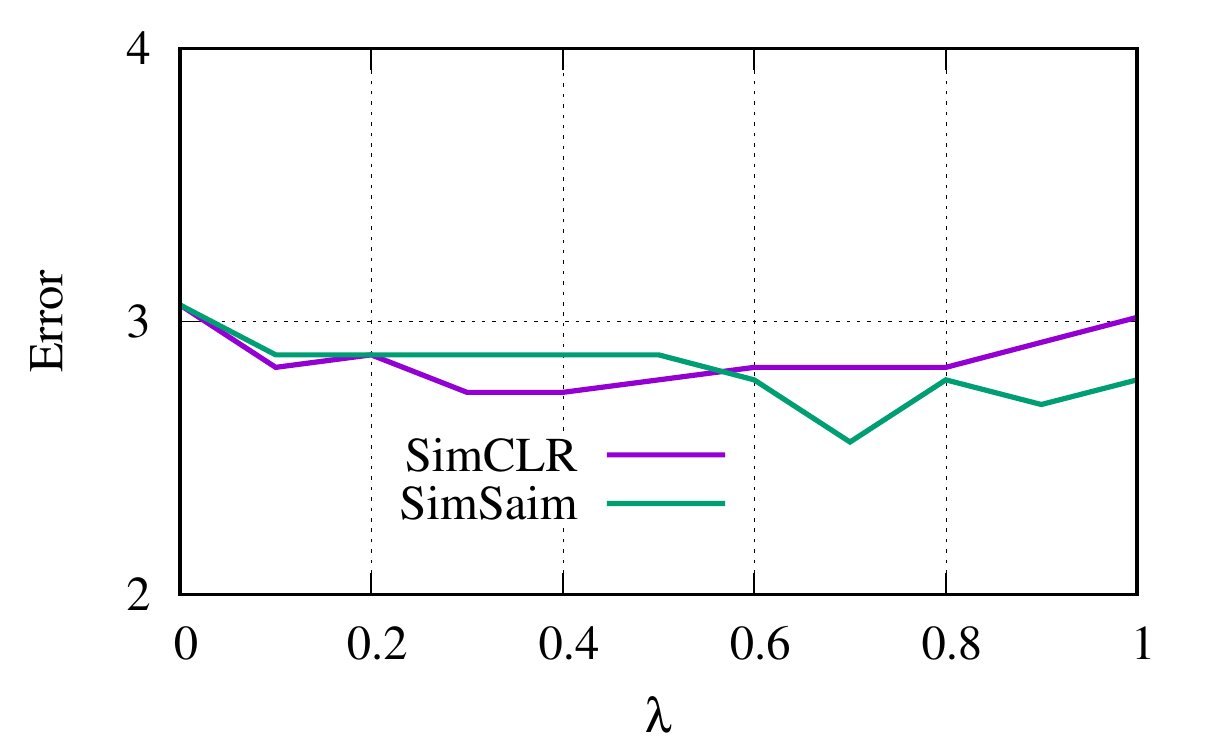}
    \caption{Parameter sensitivity study on R8 dataset.}
    \label{fig:para}
\end{figure}

In this part, we provide a parameter sensitivity study on the coefficient $\lambda$, which is used to get the trade-off between the Doc2vecC loss and contrastive losses.  We range $\lambda$ from 0 to 1 and use R8 in this set of experiments. With $\lambda=0$, \our degenerated to the original Doc2vecC model. The results are shown in  Fig.~\ref{fig:para}. It shows that our method is quite robust the the choices of $\lambda$.

% \subsection{Acknowledgements}

\end{document}